\documentclass{article}

\usepackage[final,dblblindworkshop]{neurips_2026}
\workshoptitle{Personalization, Adaptation, and Learning in Memory (PALM)}

\usepackage[utf8]{inputenc}
\usepackage[T1]{fontenc}
\usepackage{float}   
\usepackage{hyperref}
\usepackage{url}
\usepackage{booktabs}
\usepackage{amsfonts}
\usepackage{amsmath}
\usepackage{microtype}
\usepackage{xcolor}
\usepackage{tikz}
\usetikzlibrary{positioning,arrows.meta,fit,backgrounds}
\usepackage{pgfplots}
\pgfplotsset{compat=1.16}

\definecolor{litblue}{HTML}{2166AC}   
\definecolor{cauver}{HTML}{B2182C}    
\definecolor{gategray}{HTML}{6E6E6E}  

\newcommand{\ci}[1]{\,[#1]}

\floatstyle{ruled}
\newfloat{algorithm}{tb}{loa}
\floatname{algorithm}{Algorithm}
\newcounter{algln}
\newcommand{\algresetlines}{\setcounter{algln}{0}}
\newcommand{\algln}[2][0]{%
  \stepcounter{algln}%
  \noindent\makebox[1.4em][r]{{\scriptsize\textcolor{gategray}{\arabic{algln}}}}%
  \hspace{0.8em}\hspace*{#1em}#2\par}
\newcommand{\algcmt}[1]{\hfill{\footnotesize\textcolor{gategray}{$\triangleright$ #1}}}

\pgfplotsset{
  palmplot/.style={
    ymin=0, ymax=100,
    ytick={0,25,50,75,100},
    ymajorgrids,
    grid style={line width=0.4pt, draw=black!12},
    axis line style={line width=0.5pt, draw=black!45},
    tick style={draw=black!45},
    ticklabel style={font=\footnotesize},
    xticklabel style={font=\footnotesize, align=center},
    xlabel style={font=\footnotesize},
    ylabel style={font=\footnotesize},
    every axis plot/.append style={line width=1.1pt},
    clip=false,
  }
}

\title{AgentMemGate: Addressing Speculation Contamination in Conversational Assistant Memory}

\author{%
  Chirag Sharma \quad Benjamin Fowlersmith \quad Karime Maamari\\
  Distyl AI\\
  \texttt{\{chirag.sharma,benjamin.fowlersmith,karime\}@distyl.ai}%
}

\begin{document}

\maketitle

\begin{abstract}
Conversational AI assistants with long-term memory extract facts from user messages into a store consulted in later conversations. A stated plan can enter that store as fact: a user who might move to Seattle may be recorded as already living there. We call this \textit{speculation contamination}. Final-state memory benchmarks miss this error because they do not probe intermediate state and include few unresolved speculations. We present AgentMemGate, a write-time gate for profile-store memory that classifies extracted statements as speculation, completed event, correction, or other. Speculations remain outside memory, with conditions governing later promotion or deletion. We also contribute a dataset of multi-session conversations in which plans are confirmed, abandoned, or left unresolved. On our 147-conversation held-out set, Mem0 and Graphiti assert unresolved plans as current state for 35.2\% and 27.3\% of pending plans. On the core benchmark, AgentMemGate eliminates all observed contamination relative to the identical ungated pipeline (87.5\% to zero for the most exposed extraction style) and raises task accuracy from 65\% to 95\%. On the harder held-out set, gated contamination is 3.4--5.7\% and task accuracy rises by 9--13 percentage points. Our analysis identifies field matching as the main remaining bottleneck: realistic speculations often match no profile field and never reach the gate. We release our datasets, prompts, and evaluation code.
\end{abstract}

\section{Introduction}
\label{sec:intro}

Consider a user who tells an assistant: ``I've been thinking about maybe moving to Seattle, my dad lives there.'' Three weeks later, the user asks for a booking ``near where I live''. If the memory system wrote \texttt{current\_city = Seattle}, the assistant acts on a move that never happened. A plan has displaced a true fact. We call this failure \emph{speculation contamination}: a speculative statement stored as current fact.

Final-state benchmarks do not catch this because they score only after outcomes arrive. Few conversations leave speculations unresolved, and explicit confirmations can repair the store before the final question. Existing systems improve extraction, deduplication, and temporal structure but do not defer writes pending later evidence. On 147 held-out plan-to-outcome conversations, Mem0 \citep{chhikara2025mem0} answered 78.9\% of final questions correctly yet asserted never-true plans as current state for 35.2\% of pending plans. Zep Graphiti \citep{rasmussen2025zep} showed 48.3\% final accuracy and 27.3\% pending contamination. Final-answer accuracy and pending-window correctness are distinct capabilities. We define, measure, and address the latter.\footnote{Code and datasets: \url{https://github.com/DistylAI/agentmemgate-artifacts/}.}

A \emph{profile store} is the assistant's persistent set of user attributes, such as home city or employer, with one current value per field. An \emph{at-risk plan} is a stated plan that never came true, whether explicitly abandoned or still unresolved. \emph{Contamination} (i.e. speculation contamination) occurs when an at-risk plan is stored as a field's current value; we measure its rate over at-risk plans. The \emph{pending window} comprises the conversation turns after a plan is stated and before its outcome is known. The \emph{gate} is the routing step that classifies each extracted statement and decides whether to write it to the profile store. \emph{Staging} means holding a speculative statement outside the profile store with resolution conditions attached and \emph{promotion} means moving a staged item into the profile store once the user confirms that the plan happened.

We run our experiments with two extraction styles modeled on existing framework prompts: a \emph{literal extractor} that records stated values directly and a \emph{cautious extractor} that explicitly records plans as plans (Section~\ref{sec:setup}).

Our key motivation is that a speculation is not an unwarranted extraction but a fact that is not yet true. It should therefore be deferred, not filtered: hold it with resolution conditions, promote it on confirmation, and discard it on abandonment.

Our contributions:
\begin{enumerate}
  \item \textbf{An evaluation protocol that scores memory during the interaction, with a plan-to-outcome conversation dataset built for it.} The protocol reads the store within the pending window. Each conversation tracks a plan to confirmation, abandonment, or no resolution. We also evaluate on controlled-difficulty splits that vary hedge strength and confirmation wording one at a time (Section~\ref{sec:datasets}).
  \item \textbf{AgentMemGate, a write-time gate that lowers observed contamination while improving final task performance on our benchmarks.} It classifies extracted statements by evidence type and holds speculations outside the profile store until confirmation. On the core benchmark, gating reduces literal-extractor contamination from 87.5\% to zero while raising task accuracy from 65\% to 95\% (Sections~\ref{sec:approach} and~\ref{sec:gating}).
  \item \textbf{An analysis locating the remaining bottleneck at field matching.} Realistic plans often match no profile field and never reach the gate. Once staged, promotion is at parity across extraction styles. We also identify a tradeoff between looser field matching and containment of classification errors (Section~\ref{sec:coverage}).
\end{enumerate}

\begin{figure}[t]
  \centering
  \begin{tikzpicture}[scale=0.98, transform shape,
      box/.style={draw=black!55, fill=black!3, rounded corners=2pt,
                  minimum height=0.82cm, align=center, font=\small, inner sep=3pt},
      shared/.style={box},
      bad/.style={box, draw=cauver!80!black, fill=cauver!10},
      good/.style={box, draw=litblue!75!black, fill=litblue!11},
      note/.style={font=\scriptsize, text=black!65, align=center},
      panel/.style={rounded corners=4pt, inner sep=5pt},
      arr/.style={-{Stealth[length=2.1mm]}, line width=0.75pt, draw=black!60},
      badarr/.style={arr, draw=cauver!85!black},
      goodarr/.style={arr, draw=litblue!75!black}]

    \node[shared, text width=2.25cm] (msg) at (0,1.25)
      {``I might move\\to Seattle''};
    \node[shared, text width=1.85cm] (ext) at (2.75,1.25)
      {extract\\{\scriptsize home city = Seattle}};
    \node[shared, text width=1.85cm] (match) at (5.25,1.25)
      {match field\\{\scriptsize home city}};

    \node[bad, text width=1.75cm] (direct) at (7.85,1.25)
      {write directly};
    \node[bad, text width=2.15cm] (badstore) at (10.95,1.25)
      {profile store\\{\scriptsize current home city}};

    \node[good, text width=1.75cm] (cls) at (6.75,-0.55)
      {classify\\evidence};
    \node[good, text width=1.25cm] (gate) at (8.90,-0.55)
      {gate};
    \node[good, text width=2.10cm] (goodstore) at (11.65,-0.55)
      {profile store\\{\scriptsize current home city}};
    \node[good, dashed, text width=1.9cm] (stage) at (8.90,-1.75)
      {stage\\{\scriptsize Seattle (pending)}};

    \draw[arr] (msg) -- (ext);
    \draw[arr] (ext) -- (match);
    \draw[dashed, draw=gategray, line width=0.65pt, -{Stealth[length=1.8mm]}]
      (match.north) -- ++(0,0.55) node[note, above] {no field: drop};

    \draw[badarr] (match) -- (direct);
    \draw[badarr] (direct) -- (badstore);

    \draw[goodarr] (match.south) |- (cls.west);
    \draw[goodarr] (cls) -- (gate);
    \draw[goodarr] (gate.east) --
      node[note, above=2pt] {commit}
      (goodstore.west);
    \draw[goodarr] (gate.south) -- node[note, right] {speculation} (stage.north);
    \draw[goodarr, dashed] (stage.east) to[out=5,in=-105]
      node[note, below, sloped] {confirm: promote} (goodstore.south);
    \draw[goodarr, dashed]
      (stage.west) -- ++(-0.95,0) node[note, above, pos=0.5] {discard};

    \begin{scope}[on background layer]
      \node[panel, draw=cauver!55, fill=cauver!4, fit=(direct) (badstore),
            label={[font=\footnotesize\bfseries, text=cauver!80!black]above:GATE OFF}] {};
      \node[panel, draw=litblue!55, fill=litblue!4, fit=(cls) (gate) (stage) (goodstore),
            label={[font=\footnotesize\bfseries, text=litblue!75!black]above:AGENTMEMGATE}] {};
    \end{scope}
  \end{tikzpicture}
  \caption{Shared extraction and field matching feed two policies: the ungated pipeline writes speculation as current, while AgentMemGate stages it until confirmation or abandonment. With literal extraction on the core benchmark, gating takes pending contamination from 87.5\% to 0\% and task accuracy from 65\% to 95\%.}
  \label{fig:pipeline}
\end{figure}
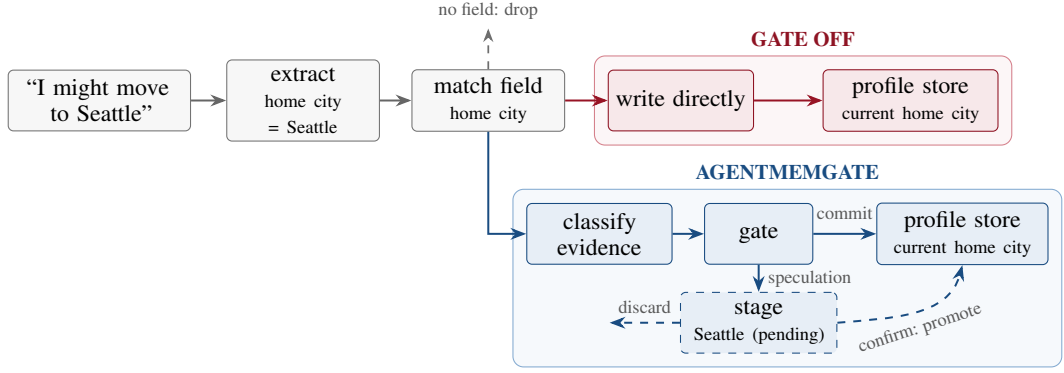

\section{Related work}
\label{sec:related}

\textbf{Memory architectures.} Long-term assistant memory increasingly manages changing persistent state rather than only retaining history. MemGPT \citep{packer2023memgpt} uses memory tiers, while Mem0 \citep{chhikara2025mem0} and Zep Graphiti \citep{rasmussen2025zep} extract structured memories from conversation. Graphiti's temporal validity intervals inspire the model for our profile store. Reflective and experience-based methods revisit memories after feedback \citep{tan2025reflective,xiong2025experience}, and Cordon \citep{chen2026cordon} gives agents transactional operations with commit conditions.

\textbf{Write-time control.} MemRouter \citep{hu2026memrouter} learns whether to store a turn, and ConsistencyGate \citep{zhang2026consistencygate} admits a fact only when repeated queries agree it is supported. Both filter writes but neither defers a fact for later resolution (Section~\ref{sec:intro}). Longitudinal-risk and memory-poisoning studies likewise place guardrails at write time \citep{altawaha2026remembering,dash2026poisoning}. These methods improve memory organization, revision, and protection but leave open when statements should enter memory.

\textbf{Memory evaluation.} LongMemEval \citep{wu2024longmemeval} and LoCoMo \citep{maharana2024locomo} test recall and reasoning across multi-session conversations; we reuse them in auxiliary evaluations (Sections~\ref{sec:setup} and~\ref{sec:results}). Change-focused benchmarks test replacement of outdated or superseded facts \citep{bae2022keep,jiang2025personamem,chao2026stale,patel2026supersede,li2026horizonbench,hu2026evermembench,chen2026esmemeval}, while TriggerBench \citep{zhang2026triggerbench} tests execution of intentions when future conditions hold. HaluMem \citep{chen2025halumem} traces hallucinations across extraction, updating, and answering. It targets fabricated or conflicting content, whereas we target faithful statements that are not yet true and score them against later outcomes. Temporally structured memories \citep{sen2026chronos,li2026timem} organize when events happened at read time; our gate decides whether they have happened at write time. Existing end-to-end or final-state scoring does not expose intermediate-state errors, so we inspect the store before plan outcomes are known.

\textbf{Linguistic factuality and belief revision.} Hedge detection and event-factuality research characterize how language marks uncertainty and whether an event occurred \citep{farkas2010conll,sauri2009factbank}, motivating our distinction between speculation and completed events. Our separation of completed events from corrections also parallels belief-revision theory's distinction between incorporating a genuine change in the world and correcting a record that was already wrong \citep{alchourron1985agm,katsuno1991difference}.

\section{Approach}
\label{sec:approach}

This section defines the profile store and the pipeline stages: extraction, field matching, classification, routing, and resolution. Algorithm~\ref{alg:turn} formalizes the flow.

A profile store contains user fields (e.g.\ home city, grocery store, employer). A field is a slot, and a fact is a proposed value for a slot that can be read later. Our baseline extracts candidate facts, matches them to fields, and writes matches. This is a controlled version of the Mem0 and Graphiti ingestion pattern of comparing candidates with stored content before writing \citep{chhikara2025mem0,rasmussen2025zep}. We fix fields in advance to measure write behavior. AgentMemGate classifies each match and either writes or holds it; later evidence promotes or discards held items (Figure~\ref{fig:pipeline}).

\paragraph{Formal store definition.} An entity $e \in E$ is who a fact concerns, and an attribute $a \in A$ is its tracked property; together they form a key $k=(e,a)$. For example, \texttt{(user, grocery\_store)} tracks which store the user visits. The profile store $P$ is a partial map from keys $K=E\times A$ to values in $V$. For each key, it stores a current value and historical values with validity intervals $[t_{\text{begin}},t_{\text{end}})$. A correction marks an old value's interval as never valid and reads return the current value. The store makes no cross-field inference: a field changes only through its own commits. Figure~\ref{fig:store} in Appendix~\ref{app:examples} illustrates the layout.

\paragraph{Extraction.} At each user turn $u_t$ with context $C_t$, the \emph{extractor} proposes tuples $(e, a, v) \in E \times A \times V$; it does not label statement types or track outcomes. We model the extractor as a model call over the current user message, timestamp, and conversation history.

\paragraph{Field matching.} The field set $F\subseteq K$ is fixed before each evaluation. $\mathrm{Match}((e,a,v),F)$ is a model call that returns the matching field $k$ for candidate $(e, a)$ or none. The model sees the tuple and each frozen field's name and description, mapping only to the same single-valued semantic slot; the value clarifies the attribute. If no match occurs, nothing is written or staged. Because each conversation has one registered field in our evaluations, the matcher accepts or declines rather than selects among fields. Note that this step is critical for end-to-end performance since a failed match never reaches AgentMemGate (Section~\ref{sec:coverage}).

\paragraph{Classification and routing.} A separate \emph{classifier} model call sees the statement's source span $s$ and up to two preceding turns, returning a distribution $q(c\mid s,C)$ over evidence types and then routing by argmax. \emph{Speculation} describes a possible future change and is held outside the store. A \emph{completed event} reports a change that happened and commits immediately. A \emph{correction} replaces a wrong value and marks it as never valid. Other observations also commit immediately. With AgentMemGate off, every matched tuple commits regardless of class.

\paragraph{Staged items and resolution.} A \emph{staged item} $\sigma=(k,v,\varphi)\in H$ carries two natural-language resolution predicates generated at hold time to avoid later re-extraction. A separate \emph{checker} model call evaluates each later user turn (example in Appendix~\ref{app:examples}): $\varphi_{\text{confirm}}(u_t)$ promotes $v$ to $P(k)$, while $\varphi_{\text{abandon}}(u_t)$ discards it. Each item therefore ends up promoted, discarded, or pending and remains invisible to reads while staged. Algorithm~\ref{alg:turn} gives the procedure.

\begin{algorithm}[t]
  \caption{A user turn under AgentMemGate. $P$ = profile store, $F$ = store fields, $H$ = staged items.}
  \label{alg:turn}
  \begin{minipage}{\linewidth}
  \small
  \algresetlines
  \algln{\textbf{procedure} \textsc{ProcessTurn}$(u_t, C_t, P, H)$}
  \algln[1.2]{$T \leftarrow \textsc{Extract}(u_t, C_t)$ \algcmt{tuples $(e, a, v)$}}
  \algln[1.2]{\textbf{for} $(e, a, v)$ \textbf{in} $T$ \textbf{do}}
  \algln[2.4]{$k \leftarrow \textsc{Match}((e, a, v), F)$ \algcmt{matching field or none}}
  \algln[2.4]{\textbf{if} $k = \text{none}$ \textbf{then} $k \leftarrow \textsc{RetryPlanFrame}((e, a, v), F)$ \algcmt{gated cautious configuration (\S\ref{sec:setup})}}
  \algln[2.4]{\textbf{if} $k = \text{none}$ \textbf{then} \textbf{continue} \algcmt{dropped before the gate}}
  \algln[2.4]{$c \leftarrow \arg\max_c\, q(c \mid s(e, a, v), C_t)$ \algcmt{evidence types (\S\ref{sec:approach})}}
  \algln[2.4]{\textbf{if} gate off \textbf{or} $c \neq \text{speculation}$ \textbf{then}}
  \algln[3.6]{$P(k) \leftarrow v$ \algcmt{commit; corrections invalidate old interval}}
  \algln[2.4]{\textbf{else}}
  \algln[3.6]{$H \leftarrow H \cup \{(k, v, \varphi)\}$ \algcmt{stage; invisible to reads}}
  \algln[1.2]{\textbf{for} $\sigma = (k, v, \varphi)$ \textbf{in} $H$ \textbf{do} \algcmt{check staged conditions}}
  \algln[2.4]{\textbf{if} $\varphi_{\text{confirm}}(u_t)$ \textbf{then} $P(k) \leftarrow v$;\; $H \leftarrow H \setminus \{\sigma\}$ \algcmt{promote}}
  \algln[2.4]{\textbf{else if} $\varphi_{\text{abandon}}(u_t)$ \textbf{then} $H \leftarrow H \setminus \{\sigma\}$ \algcmt{discard}}
  \end{minipage}
\end{algorithm}

A correct final value can arrive via a promotion or a direct commit of the user's confirmation. For example, ``I bought them yesterday'' is a completed event that any system, gated or not, commits. We call these \emph{confirmation-commits}; they can lead to correct end states in ungated systems despite contaminated intermediate memory.

\section{Experimental setup}
\label{sec:setup}

\paragraph{Pipelines.} We evaluate four versions of the pipeline: the two extractors of Section~\ref{sec:intro} crossed with AgentMemGate on/off. The literal extractor records stated values directly, e.g.\ \texttt{(user, grocery\_store, ``Trader Joe's'')}, and the cautious extractor records plans as plans, e.g.\ \texttt{(user, grocery\_plan, ``considering Trader Joe's'')}. Their prompts are derived from Zep Graphiti's and Mem0's published extraction prompts, adapted to our tuple interface while preserving each product's extraction policy. All four evaluations share the same profile store, probes, and judge. Since plan-framed attributes in the cautious configuration are more susceptible to failing field matching, we introduce a single deterministic retry mechanism: plan-related phrases are RegEx-matched and removed and the remainder is resubmitted to the matcher (e.g.\ \texttt{grocery\_plan $\rightarrow$ grocery $\rightarrow$ grocery\_store}). The original speculative statement still reaches the classifier, so it is held rather than committed. An ablation confirms that the retry is not doing the gate's work (Section~\ref{sec:coverage}). Extraction, field matching, classification, and resolution checking are each single calls to the pinned LLM below. Routing and the store itself are deterministic.

\paragraph{Evaluation protocol.} We report three measures. \textbf{Pending-window contamination} is the fraction of at-risk plans stored as a field's current value while the plan is open. \textbf{At-end contamination} is the same fraction at conversation end. \textbf{Task accuracy} is measured by one probe per conversation (``what does the user currently use?'') answered from the final store and scored against ground truth by the fixed judge below.

\paragraph{Baselines and controls.} The \emph{ungated baseline} is the same pipeline with the gate disabled, in both extraction styles. The \emph{prompt-only baseline} is the cautious ungated configuration. It tests whether phrasing plans as plans is sufficient without a staged write policy. The \emph{plan-typed baseline} writes each speculative value to the profile store as a labeled plan and hides it from reads, testing whether read-time masking can replace staging and resolution (Appendix~\ref{app:plantyped}). Two \emph{controls} test the over-holding cost of gating: already-completed changes must commit immediately, and explicit corrections must apply (Section~\ref{sec:gating}).

\paragraph{Product evaluations.} We also run black-box Mem0 and Graphiti evaluations for motivation and external context, but they are not causal baselines for our component-level comparison. We ran each product end to end with its own extraction, storage, and retrieval, configured to use the same pinned model as our systems. Each of the 147 held-out conversations was ingested session by session. After every session, the product was asked for the user's current value, with answers scored by the same equivalence judge. A plan counts as exposed if any answer before its outcome asserts it as the user's current state. Answers worded as plans do not count (measurement detail in Appendix~\ref{app:products}). Although the literal extractor follows Graphiti's published extraction prompt, end-to-end pipeline behavior may vary because the product adds retrieval, deduplication, and answer-time machinery around extraction.

\paragraph{Models and scoring.} All system components run on a pinned model, \texttt{gpt-5.5-2026-04-23}, with fixed prompts. Scoring uses one fixed equivalence judge (the same pinned model) that compares a stored or answered value with the labeled value. Judge validation has two layers. A model from a different family re-judged 150 pairs, sampled to over-represent borderline cases, and agreed with the judge on 90.7\% of cases. Furthermore, a blind human audit on a sample of 54 pairs agreed with the judge on 87.0\%\ci{75.6, 93.6} ($\kappa = 0.74$). All disagreements fell on borderline pairs. In five of the seven, the judge was the stricter of the two, so judge error tends to lower the reported accuracies, not raise them. Proportions carry 95\% Wilson confidence intervals (one observation per conversation), and paired differences use a cluster bootstrap.

\paragraph{Classifier validation.} We evaluated the gate classifier on 163 naturally occurring statements from LongMemEval. The reference labels for evidence type are model-annotated with human audit: three model votes under a fixed rubric produce each label, and a model from a different family relabeled all 163 statements, agreeing on 92.0\%. A blind human audit of 40 statements sampled from the items where the two model families agreed matched the reference label on 87.5\%\ci{73.9, 94.5}. When restricted to the binary commit-versus-hold decision used for routing, agreement was 92.5\%\ci{80.1, 97.4}. The 13 statements the model families disputed were audited separately and remain contested. The classifier prompt was fixed before any reference label was produced, but its class taxonomy was developed on the same corpus. We then tested whether the classifier gives higher update scores to statements that should update the profile. Its AUROC is 0.93\ci{0.89, 0.96}, where 1.0 is perfect and 0.5 is chance. Two simpler prompts perform worse on the same test: asking whether the fact is true now scores 0.69, and asking whether it should be stored scores 0.62. Appendix~\ref{app:classifier} gives detailed results and the repeated-run analysis.

\paragraph{Overhead.} Relative to the gate-off baseline, AgentMemGate adds one classifier call per matched
fact and one checker call per staged item per future user turn.

\section{Datasets}
\label{sec:datasets}

\begin{table}[t]
  \caption{The plan-to-outcome dataset: one construction family, three splits. $n$ counts conversations.}
  \label{tab:datasets}
  \centering
  \footnotesize
  \begin{tabular}{@{}p{2.9cm}rp{5.9cm}p{2.9cm}@{}}
    \toprule
    Split & $n$ & What varies & Role \\
    \midrule
    Core benchmark & 60 & Plan outcome: 20 plans come true, 20 are abandoned, 20 stay open & Primary benchmark (\S\ref{sec:gating}) \\
    Held-out set & 147 & Shorter sessions, 44 in each plan outcome bucket + 15 completed-change controls, values are long free-text phrases & Held-out replication (\S\ref{sec:gating}) \\
    Controlled-difficulty splits & 179 & One factor at a time: hedge strength across 5 levels, confirmation wording across 4 & Difficulty curves (\S\ref{sec:hedge}--\ref{sec:coverage}) \\
    \bottomrule
  \end{tabular}
\end{table}

In an audit of existing public data, we used an LLM to extract 194 naturally occurring hedge statements from LongMemEval and LoCoMo. Only 16 at-risk plans could be measured through an outcome, so we built our own plan-to-outcome dataset in Table~\ref{tab:datasets}. All three splits share the same overall structure: 2--3 dated sessions in which the user states a current value, hedges a switch, and returns later, with each plan labeled by its outcome (came true, abandoned, or still open). 

The core benchmark split balances the three plan outcomes. The held-out set is a larger set of conversations with the same structure, but with longer free-text extracted values for higher difficulty. This dataset is used only for final evaluation. The controlled-difficulty splits vary one factor at a time: four different levels of plan wording (strongly hedged, weakly hedged, confident plan without a date, scheduled plan with date), plus a completed-change control level, and four different variants of confirmation wording (verbatim, paraphrase, oblique, situational). The completed-change control examples in the held-out set and controlled-difficulty splits are cases where the user reports a change that has \emph{already happened} instead of speculating a change; this is to measure potential over-holding. An example arc, the classifier prompt, and field-set construction are in Appendix~\ref{app:examples}.

\paragraph{Validation.} Validation combined blind model judging with a separate human review. The core benchmark passed an LLM-judge review that checks that conversations read as natural human chat and that each plan statement is motivated by its context, followed by an item-level human review. For the controlled-difficulty splits, blind judges recovered the intended difficulty level (hedge strength or confirmation wording) in 89\% of conversations. The held-out split passed the same validation.

\paragraph{Provenance.} All three splits are LLM-generated and reviewed by the authors. Generators are GPT-family, the same vendor family as the evaluated systems, so the held-out split is held-out but not generator-independent. The validation judges are Claude-family, a different model family from both.

\section{Results}
\label{sec:results}

\subsection{AgentMemGate removes pending-window contamination at little task cost}
\label{sec:gating}

\begin{table}[t]
  \caption{Contamination: percent of at-risk plans (abandoned + still open) stored as a current fact. Lower is better. 95\% Wilson CIs. Measures and baseline: the evaluation protocol in Section~\ref{sec:setup}.}
  \label{tab:contamination}
  \centering
  \footnotesize
  \setlength{\tabcolsep}{3.6pt}
  \begin{tabular}{@{}lllll@{}}
    \toprule
    Measurement (source, $n$) & Cautious ungated & Cautious gated & Literal ungated & Literal gated \\
    \midrule
    While pending (core benchmark, 40 plans) & 2.5\%\ci{0.4, 12.9} & 0\%\ci{0, 8.8} & 87.5\%\ci{73.9, 94.5} & 0\%\ci{0, 8.8} \\
    At end (core benchmark, 40 plans) & 2.5\%\ci{0.4, 12.9} & 0\%\ci{0, 8.8} & 47.5\%\ci{32.9, 62.5} & 0\%\ci{0, 8.8} \\
    At end (held-out, 88 plans) & 10.2\%\ci{5.5, 18.3} & 5.7\%\ci{2.5, 12.6} & 23.9\%\ci{16.2, 33.7} & 3.4\%\ci{1.2, 9.5} \\
    \bottomrule
  \end{tabular}
\end{table}

\textbf{Ungated, the literal extractor stores most pending plans as facts.} While a plan is open, 87.5\%\ci{73.9, 94.5} of at-risk plans sit in the store as current values, and much of that error survives to conversation end (Table~\ref{tab:contamination}). Confirmation-commits repair most abandoned plans only after the user states the outcome, and open plans get no such message and stay wrong. The ungated cautious extractor mostly protects against contamination via plan phrasing: at-risk plan contamination is only 2.5\%\ci{0.4, 12.9}.

\textbf{Gated, low to no contamination is observed.} For both literal and cautious extractor styles, we measure 0 (upper bound 8.8\%) pending-window and at-end contamination with the gated configuration on the core benchmark. Section~\ref{sec:hedge} attributes the cautious-gated zeros. On the held-out set, at-end contamination rises slightly to 5.7\%\ci{2.5, 12.6} in the cautious gated case and 3.4\%\ci{1.2, 9.5} in the literal gated case (Table~\ref{tab:contamination}, bottom row).

\begin{table}[t]
  \caption{Task accuracy: one probe per conversation answered from the final store (evaluation protocol in Section~\ref{sec:setup}).}
  \label{tab:task}
  \centering
  \footnotesize
  \setlength{\tabcolsep}{3.0pt}
  \begin{tabular}{@{}lllll@{}}
    \toprule
    Measurement (source, $n$) & Cautious ungated & Cautious gated & Literal ungated & Literal gated \\
    \midrule
    Core benchmark (60 questions) & 88.3\%\ci{77.8, 94.2} & 90.0\%\ci{79.9, 95.3} & 65.0\%\ci{52.4, 75.8} & 95.0\%\ci{86.3, 98.3} \\
    Held-out set (147 conversations) & 53.1\%\ci{45.0, 60.9} & 62.6\%\ci{54.5, 70.0} & 53.7\%\ci{45.7, 61.6} & 66.7\%\ci{58.7, 73.8} \\
    \bottomrule
  \end{tabular}
\end{table}

\textbf{The gate does not trade accuracy for correctness.} It moves the literal style from worst to best on the core benchmark (Table~\ref{tab:task}), which indicates that contamination caused many of the ungated errors.  On the held-out set, the gate improves task accuracy for both extractors ($+$12.9 and $+$9.5 points), while the core-benchmark cautious gain (88.3\% to 90.0\%) is within noise. We also test 161 questions from LongMemEval and LoCoMo whose answers depend on a fact that changes during the conversation. We run each question with the gate on and off. The paired gate-on minus gate-off difference is $+0.0$ points\ci{$-$4.8, $+$4.3} for the cautious style and $+9.9$ points\ci{$+$3.1, $+$17.8} for the literal style (95\% cluster-bootstrap intervals). Two control tests cover non-speculative updates. Across 240 runs containing an explicit correction, the gate-on and gate-off systems produce the same result. On the 15 completed-change controls from the held-out set, each gated configuration incorrectly holds one of the changes instead of storing immediately, i.e., 6.7\% over-holding rate (Appendix~\ref{app:heldout}).

\textbf{The contrasts hold on the held-out set.} At-end contamination remains the worst for the literal ungated configuration and the gate cuts contamination and improves task accuracy in both styles (Table \ref{tab:contamination}, bottom row). We do not report pending contamination for the held-out set because its conversations place the outcome immediately after the plan, leaving no pending window to measure. Note that absolute contamination levels may be below the core benchmark numbers because of the longer free-text extracted values in this dataset. The full table and two measurement caveats are in Appendix~\ref{app:heldout}.

We also tested a read-time alternative that commits all speculations but explicitly marks them as plans, and then hides them from reads (Appendix~\ref{app:plantyped}). With appropriate plan confirmation, it matches the gate's contamination and end-state correctness. Without it, literal came-true correctness falls to 5.0\% (Appendix~\ref{app:plantyped}). However, it leaves 85--90\% of plan entries unresolved at conversation end.

A full rerun on a second model family (\texttt{claude-opus-4-8}) reproduces 15 of 18 directional findings, with exceptions favoring the gate or within noise (Appendix~\ref{app:secondfamily}).

Overhead is small: on the core benchmark, the gate adds 0.8 model calls per user message under the literal style and 0.4 under the cautious style, excluding the retry's occasional extra matching call.

\subsection{Failures concentrate where plans sound settled and confirmations are indirect}
\label{sec:hedge}

The hedge-strength split (Table~\ref{tab:datasets}, controlled-difficulty splits) realizes the same speculation at five levels. At the four open-plan levels (strongly hedged, weakly hedged, confident plan without a date, and scheduled plan with a date), the plan stays open (20 conversations each), so any stored new value is contamination. The fifth level is a completed-change control (19 conversations). Note that a stronger hedge indicates weaker commitment and is therefore a less settled plan (e.g.\ ``I'm torn between staying with Walgreens and switching to CVS''). The complete set of results is in Table~\ref{tab:hedgeexact}, Appendix~\ref{app:exact}.

\begin{figure}[t]
  \centering
  \begin{tikzpicture}
    \begin{axis}[
      palmplot,
      ymin=-4,
      width=0.72\linewidth, height=5.6cm,
      symbolic x coords={SH, WH, CP, SD},
      xtick=data,
      xticklabels={strongly hedged, weakly hedged, {confident plan\\(no date)}, {scheduled\\(with date)}},
      xlabel={plan wording (more settled $\rightarrow$)},
      ylabel={\% of at-risk plans stored as current fact},
      enlarge x limits=0.07,
      legend style={
        font=\footnotesize, draw=none, fill=none,
        at={(0.5,1.02)}, anchor=south, legend columns=3,
        /tikz/every even column/.append style={column sep=0.55cm},
        cells={anchor=west},
      },
      legend cell align=left,
    ]
      \addplot[black!60, dashed, line width=1.6pt, mark=none]
        coordinates {(SH,0) (WH,0) (CP,0) (SD,0)};
      \addlegendentry{gated (both styles)}
      \addplot[cauver, mark=square*, mark size=2.1pt, mark options={fill=cauver, draw=white, line width=0.5pt}]
        coordinates {(SH,0) (WH,0) (CP,10) (SD,20)};
      \addlegendentry{cautious ungated}
      \addplot[litblue, mark=*, mark size=2.2pt, mark options={fill=litblue, draw=white, line width=0.5pt}]
        coordinates {(SH,85) (WH,75) (CP,85) (SD,90)};
      \addlegendentry{literal ungated}
    \end{axis}
  \end{tikzpicture}
  \caption{At-end contamination by hedge strength ($n$=20 per level). Ungated, the literal style remains high throughout, while the cautious style rises as plans sound more settled. Both gated styles remain at 0/20 (Wilson upper bound 16.1\%).}
  \label{fig:hedge}
\end{figure}
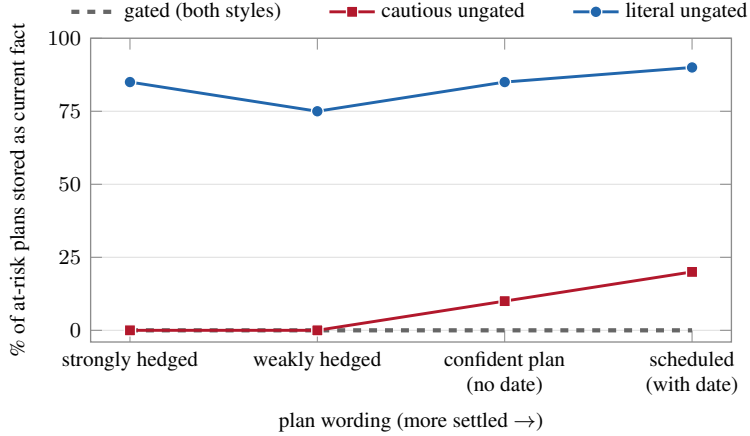

\textbf{Ungated failure concentrates where plans sound most settled.} The ungated literal pipeline almost saturates contamination, storing hedges as facts at every level. The ungated cautious style maintains 0 contamination at both hedged levels but contamination levels increase as plan certainty increases (Figure~\ref{fig:hedge}). Its protection tracks wording, not evidence, with nothing tracking resolution, so by itself it is particularly sensitive to choice of wording.

\textbf{The literal extractor isolates the effect of the gate.} In the literal style, 76 of 80 hedges match a field and reach the gate. The gate then stages all 76, i.e., no hedged value is stored as a current fact. In the cautious extractor, only 40 of 80 hedges reach the gate. For 22 missing hedges, the extractor fails to extract the new state, and the other 18 fail field matching. Again, the gate stages all 40 hedges that it receives. Thus, the cautious extractor's conservative extraction has a real field-matching cost even if raw contamination rate is low. The literal result measures the gate more directly.

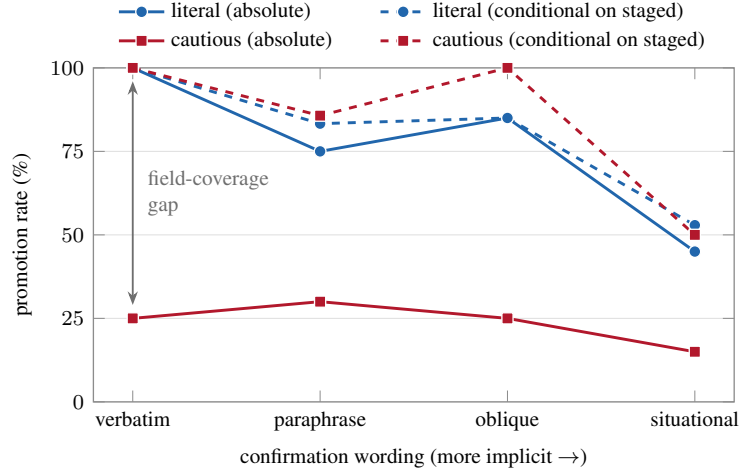
\begin{figure}[t]
  \centering
  \begin{tikzpicture}
    \begin{axis}[
      palmplot,
      width=0.72\linewidth, height=6.0cm,
      symbolic x coords={V, P, O, S},
      xtick=data,
      xticklabels={verbatim, paraphrase, oblique, situational},
      xlabel={confirmation wording (more implicit $\rightarrow$)},
      ylabel={promotion rate (\%)},
      enlarge x limits=0.07,
      legend style={
        font=\footnotesize, draw=none, fill=none,
        at={(0.5,1.02)}, anchor=south, legend columns=2,
        /tikz/every even column/.append style={column sep=0.55cm},
        cells={anchor=west},
      },
      legend cell align=left,
    ]
      \addplot[litblue, mark=*, mark size=2.2pt, mark options={fill=litblue, draw=white, line width=0.5pt}]
        coordinates {(V,100) (P,75) (O,85) (S,45)};
      \addlegendentry{literal (absolute)}
      \addplot[litblue, dashed, mark=*, mark size=2.2pt, mark options={solid, fill=litblue, draw=white, line width=0.5pt}]
        coordinates {(V,100) (P,83.3) (O,85) (S,52.9)};
      \addlegendentry{literal (conditional on staged)}
      \addplot[cauver, mark=square*, mark size=2.1pt, mark options={fill=cauver, draw=white, line width=0.5pt}]
        coordinates {(V,25) (P,30) (O,25) (S,15)};
      \addlegendentry{cautious (absolute)}
      \addplot[cauver, dashed, mark=square*, mark size=2.1pt, mark options={solid, fill=cauver, draw=white, line width=0.5pt}]
        coordinates {(V,100) (P,85.7) (O,100) (S,50)};
      \addlegendentry{cautious (conditional on staged)}
      \draw[{Stealth[length=1.8mm]}-{Stealth[length=1.8mm]}, gategray, line width=0.7pt]
        (axis cs:V,29) -- (axis cs:V,96)
        node[midway, right=2pt, font=\footnotesize, text=gategray, align=left]
        {field-coverage\\gap};
    \end{axis}
  \end{tikzpicture}
  \caption{Promotion of AgentMemGate-held plans that came true by confirmation wording ($n$=20 per level; Section~\ref{sec:datasets}). Promotion falls as wording gets more implicit. Conditional trends (dashed) are similar across styles; the split in absolute rates (solid) reflects a field coverage gap.}
  \label{fig:promotion}
\end{figure}

\textbf{Promotion is less reliable when confirmation is less direct} (Figure~\ref{fig:promotion}). Implied confirmations are exactly where confirmation-commits also fail, since the implication may not successfully get extracted as a fact. This is where AgentMemGate performance degrades. At the most implicit level of confirmation wording (situational confirmation: e.g. the user mentioned planning to do an activity in the past and now casually references already doing it), the gated literal system answers 13 of 20 task probes correctly, compared with 19 of 20 for the ungated system. The ungated system scores these because it committed the plan as a fact when stated. These came-true conversations happen to reward that premature write. The gated system holds the plan and loses cases where the implied confirmation neither promotes it nor commits on its own. Complete results are in Table~\ref{tab:promoexact}, Appendix~\ref{app:exact}. Promotion rates conditional on staging are similar across the two extraction styles but there is a wide gap in absolute promotion rates. The difference in absolute promotion therefore comes from field coverage. Section~\ref{sec:coverage} examines this gap.

\subsection{Hedged plans often match no field, so they never reach the gate}
\label{sec:coverage}

The main remaining failures occur before classification: many extracted plans do not match a profile field.

\textbf{Field matching explains the promotion gap.} Across the entire confirmation-wording split (four wording levels with 20 came-true conversations each), the cautious-gated system promotes 19 of 80 plans (23.8\%). Conditional on staging, it promotes 19 of 23 (82.6\%). The difference occurs because only 23 of the plans match a field and enter staging. The core benchmark shows the same pattern. Without the plan-frame retry (Algorithm \ref{alg:turn}, Section \ref{sec:setup}), 9 of 60 hedges reach the gate via the cautious extractor, compared with 55 of 60 for the literal extractor. The retry increases this to 31 of 60 for the cautious extractor. Once a plan is staged, promotion is similar across the two styles.

\textbf{The retry widens coverage but is safe only with the gate.} Removing the retry from the gated cautious system does not change contamination, task accuracy, or control outcomes. On the controlled-difficulty splits, it reduces staging of plans by 24.0 percentage points\ci{$-$30.7, $-$17.9} and (absolute) promotion of plans by 21.2 points\ci{$-$30.0, $-$12.5}. However, the system still reduces at-end contamination by 7.5 points\ci{$-$13.8, $-$2.5} relative to the ungated cautious system on the controlled-difficulty splits and improves held-out task accuracy by 8.2 points\ci{$+$2.7, $+$13.6}. The held-out at-end contamination difference ($-$4.5 points\ci{$-$11.4, $+$2.3}) is directionally consistent but spans zero. Adding the same retry to the ungated system has the opposite effect. Pending contamination increases by 37.5 points\ci{27.5, 48.7}, and task accuracy decreases by 15.1 points\ci{9.5, 20.7}, because more plans are extracted and stored as facts.

\textbf{Extraction succeeds, but matching often fails.} Both extractors emit tuples on every plan turn, yet the cautious extractor produces 89 unmatched tuples on held-out plan turns. Of these, 66 are represented not as explicit plans but as preferences, interests, concerns, or conditions. For example, ``might try the black messenger with the wide strap'' was extracted as \texttt{conditional\_bag\_interest}, which matched no field although the store tracked the user's main work bag. The fixed RegEx retry misses many such tuples, recovering plans in 22 of 60 core-benchmark conversations but only 11 of 132 held-out conversations.

\textbf{A more permissive matcher exposes classifier errors.} When the original matcher declines a tuple, we explored using a second LLM call to file it under any field tracking the same quantity. On a planted test suite with 60 probes of confident plans, this raises staging from 11 to 41 of 60 but also increases contamination by 5 points because the classifier commits some confidently worded plans. Thus, matching and classification involve a clear tradeoff.

\section{Discussion, limitations, and conclusion}
\label{sec:discussion}

Memory benchmarks should measure intermediate state, not only final answers. End-state scoring captures recovery but misses contamination while a plan is open, even though an assistant can act on stored state long after it was written. For values reaching the gate, the policy is close to solved; upstream coverage is now the main problem. Open schemas do not remove this risk: they can change ``plan never enters memory'' into ``plan enters as fact''.

Why use a separate memory stage instead of a stronger extraction prompt? Prompting is unstable: cautious-extractor protection weakens as plans sound more settled (Figure~\ref{fig:hedge}). A combined stage also couples extraction and routing errors, obscuring attribution and tradeoffs. AgentMemGate makes storage explicit and testable; its evidence-type classification outperforms direct truth or storage classification by $+0.24$ and $+0.31$ AUROC (Appendix~\ref{app:classifier}). Staging preserves the plan without exposing it in responses, allowing later confirmation to promote it.

Five limitations bound generalization. First, the unknown prevalence of at-risk pending plans means our constructed datasets do not establish real-world performance (Section~\ref{sec:datasets}). Their realism also needs validation despite blind judging and author review; longitudinal corpora such as REALTALK \citep{lee2025realtalk} may help. Second, small controlled splits yield wide intervals (Section~\ref{sec:gating}), motivating larger and broader datasets. Third, we do not test whether staging harms answers about plans. Fourth, each sample uses one fixed field and accept-or-reject matching, so results do not cover selection within large field sets. Finally, gate costs concentrate at implied confirmations (Section~\ref{sec:hedge}). Future work should examine confirmation wording, recovery of implied confirmations, and interactive clarification for long-pending plans. Retention limits and user control over staged content remain out of scope but matter for deployment.

Systems can mistake plans for facts while final-answer accuracy hides the error. AgentMemGate reduces contamination at low task cost through evidence classification, staging, and promotion while preserving genuine updates. Field matching remains the main bottleneck.

\bibliographystyle{plainnat}
\bibliography{refs}

\appendix

\section{Additional results and implementation details}
\label{app:appendix}

\subsection{Plan-typed commit baseline}
\label{app:plantyped}

This baseline replaces AgentMemGate routing with a read-time framework (Section~\ref{sec:gating}). It stores each speculation immediately, marks the value as a plan, and preserves the validity history. Reads ignore plan-typed values and return the most recent factual value; if none exists, they abstain. The baseline has no staging area, resolution conditions, or checker. A confirmation can therefore resolve a plan only by creating a new commit.

We test both extraction styles on confirmation wording split from the controlled-difficulty set. The same judge scores all systems. The gated and ungated systems in Section~\ref{sec:gating} provide the comparison. We report two interpretations. Under the \emph{refresh-confirm} reading, a confirmation resolves a masked plan when it commits the same value as a fact. Under the \emph{strict} reading, plans never resolve (only the confirmation statement is committed). With the strict reading, end-state correctness for came-true plans is 45/80 (56.3\%) for the cautious style and 4/80 (5.0\%) for the literal style. We report strict-reading results only at the aggregate level.

\begin{table}[htbp]
  \caption{End-state correctness for came-true plans in the plan-typed baseline under the refresh-confirm reading ($n$=20 per confirmation-wording level; controlled-difficulty set). Gated results are included for comparison.}
  \label{tab:plantyped}
  \centering
  \footnotesize
  \begin{tabular}{@{}lllll@{}}
    \toprule
    Wording level & Plan-typed cautious & Plan-typed literal & Gated cautious & Gated literal \\
    \midrule
    Verbatim & 19/20 (95.0\%) & 20/20 (100\%) & 19/20 (95.0\%) & 20/20 (100\%) \\
    Paraphrase & 16/20 (80.0\%) & 19/20 (95.0\%) & 16/20 (80.0\%) & 19/20 (95.0\%) \\
    Oblique & 15/20 (75.0\%) & 20/20 (100\%) & 15/20 (75.0\%) & 19/20 (95.0\%) \\
    Situational & 8/20 (40.0\%) & 12/20 (60.0\%) & 7/20 (35.0\%) & 13/20 (65.0\%) \\
    \bottomrule
  \end{tabular}
\end{table}

\textbf{Most literal-style successes depend on the refresh-confirm rule.} Of its 71 came-true successes, 67 are correct only under this rule. The same is true for 13 of 58 cautious-style successes. Thus, the literal baseline matches the gate mainly because equality between the plan and confirmation is treated as resolution.

\textbf{Contamination is low, but unresolved plans accumulate.} Under both readings, contamination is nearly absent. On the controlled-difficulty splits, the cautious and literal styles contaminate 0/80 and 1/80 cases, respectively, compared with 0/80 for both gated variants. The core benchmark shows the same pattern: 0/40 and 1/40, compared with 0/40. Completed changes are not the problem. The baseline commits them directly, matches the gated results (18/19 cautious and 17/19 literal), and masks none as plans. The limitation appears when plans remain open or are confirmed indirectly. At conversation end, 20--65\% of cautious-style open plans and 75--90\% of literal-style open plans are still masked. Overall, 85--90\% of plan entries remain unresolved. Even after a situational confirmation, 3/20 cautious and 7/20 literal values remain masked. The gate instead promotes or discards every staged item in these comparisons.

\textbf{Held-out results confirm that masking substitutes for resolution.} Both plan-typed styles expose only 4/88 at-risk plans as current, or 4.5\%\ci{1.8, 11.1}. This is comparable to gated contamination (5.7\% cautious and 3.4\% literal) and below the ungated rates (10.2\% and 23.9\%). However, low contamination does not mean that the baseline manages plans well. Came-true correctness is close to the gate for the cautious style (43.2\% vs.\ 47.7\%) but lower for the literal style (50.0\% vs.\ 59.1\%). When equality no longer counts as resolution, these rates fall to 43.2\% and 38.6\%. The state traces show why: by conversation end, 31.7\% of cautious-style and 32.5\% of literal-style plan commits remain current and masked. No cautious entry resolves (0 of 82 flips), and only 7 of 114 literal entries resolve. Consequently, 18.9--31.1\% of non-control conversations end with a masked value. The gate instead gives every staged item a defined end state.

\subsection{Held-out replication detail}
\label{app:heldout}

\begin{table}[htbp]
  \caption{Held-out replication (147 conversations; 132 plans: 88 at-risk, 44 came-true; 15 controls). 95\% Wilson CIs. Dashes mark measures that exist only in gated cells.}
  \label{tab:heldout}
  \centering
  \footnotesize
  \setlength{\tabcolsep}{3.6pt}
  \begin{tabular}{@{}lllll@{}}
    \toprule
    Measurement ($n$) & Cautious ungated & Cautious gated & Literal ungated & Literal gated \\
    \midrule
    Contamination at end (88) & 10.2\%\ci{5.5, 18.3} & 5.7\%\ci{2.5, 12.6} & 23.9\%\ci{16.2, 33.7} & 3.4\%\ci{1.2, 9.5} \\
    Plans staged at the plan turn (132) & --- & 33.3\%\ci{25.9, 41.7} & --- & 55.3\%\ci{46.8, 63.5} \\
    Came-true plans promoted (44) & --- & 22.7\%\ci{12.8, 37.0} & --- & 38.6\%\ci{25.7, 53.4} \\
    Abandoned plans discarded by end (44) & --- & 100\%\ci{92.0, 100} & --- & 90.9\%\ci{78.8, 96.4} \\
    Task probes correct (147) & 53.1\%\ci{45.0, 60.9} & 62.6\%\ci{54.5, 70.0} & 53.7\%\ci{45.7, 61.6} & 66.7\%\ci{58.7, 73.8} \\
    Controls committed at once (15) & 73.3\%\ci{48.0, 89.1} & 73.3\%\ci{48.0, 89.1} & 80.0\%\ci{54.8, 93.0} & 80.0\%\ci{54.8, 93.0} \\
    \bottomrule
  \end{tabular}
\end{table}

\textbf{The held-out set reproduces the main contrasts.} Both gated systems remove most contamination (Table~\ref{tab:heldout}). They discard abandoned plans reliably: 100\%
(cautious) and 90.9\% (literal). Field matching remains the bottleneck: the two styles stage 33.3\% and 55.3\% of plans and promote 22.7\% and 38.6\% of came-true plans. For the literal style, the gate improves task accuracy by 12.9 percentage points (23 conversations improve,
4 get worse).

\textbf{Immediate commit and over-holding are different control measures.} The immediate-commit rate is the same with the gate on and off (Table~\ref{tab:heldout}), so the
shortfall from 100\% arises before the gate. Over-holding asks whether a gated system stages a completed change. This happens for 1 of 15 controls (6.7\%). That item restates, in the second session, a change that already committed in the first session. The restatement does not change the immediate-commit rate; the ungated pipeline applies it as a refinement of the stored value.

\textbf{Two results differ from the core benchmark.} On the core benchmark, the gate left cautious-style task accuracy unchanged; here it improves it by 9.5 points (18 conversations
improve, 4 get worse; +8.2 points without the plan-frame retry, Section~\ref{sec:coverage}). On the core benchmark, the literal ungated system answered worst; here it ties the cautious ungated system.

\textbf{Pending-window contamination is unmeasured on the held-out set.} In the held-out set, conversations are structured so that they place the outcome of a plan immediately after the plan statement, so there is no pending window, and we report only at-end contamination (Section~\ref{sec:gating}).

\subsection{Second-model-family scorecard}
\label{app:secondfamily}

We reran all system components on \texttt{claude-opus-4-8} and used the same \texttt{gpt-5.5} judge (the main experiments use a pinned \texttt{gpt-5.5-2026-04-23} model for the components). 15 of 18 directional findings transfer immediately: the gate still suppresses contamination, the two extraction styles remain different, staging remains more common than promotion, and the graded difficulty curves remain. We measure these curves with a planted-probe analog of the controlled-difficulty factors.

The 3 differences do not reverse the main findings. First, the cautious-style task gain is $+8.3$ points for the Claude model. This agrees with the $+9.5$-point gain on the held-out set in the original evaluation and supports the finding that the gate can improve task accuracy. Second, the literal ungated system has the lowest task accuracy on Claude. It tied for the lowest accuracy in the original run, so the Claude result strengthens the same direction. Third, one control block scores 88.3\% instead of 90.0\%, a minor difference of 1.7 points.

A tangential observation is that the plan-frame retry's contamination cost appears model-sensitive: with the retry, contamination at confident plans is 28.3\% on Claude versus 15.0\% on the GPT model.

\subsection{Exact values for Figures 2 and 3}
\label{app:exact}

Tables~\ref{tab:hedgeexact} and~\ref{tab:promoexact} report the values plotted in Figures~\ref{fig:hedge} and~\ref{fig:promotion}.

\begin{table}[H]
  \caption{At-end contamination by hedge-strength level ($n$=20 per cell). Each observed zero has a 16.1\% Wilson upper bound.}
  \label{tab:hedgeexact}
  \centering
  \footnotesize
  \begin{tabular}{@{}lllll@{}}
    \toprule
    Level & Cautious ungated & Cautious gated & Literal ungated & Literal gated \\
    \midrule
    Strongly hedged & 0\% & 0\% & 85.0\% & 0\% \\
    Weakly hedged & 0\% & 0\% & 75.0\% & 0\% \\
    Confident plan, no date & 10.0\% & 0\% & 85.0\% & 0\% \\
    Scheduled, with date & 20.0\% & 0\% & 90.0\% & 0\% \\
    \bottomrule
  \end{tabular}
\end{table}

For promotion, the rate over all came-true plans includes field matching and checking. The rate among staged plans isolates the checker. A promotion is valid only when it occurs at the resolution turn.

\begin{table}[H]
  \caption{Promotion by confirmation-wording distance ($n$=20 per level). The cautious-gated checker promotes 19/23 staged plans overall (83\%).}
  \label{tab:promoexact}
  \centering
  \footnotesize
  \begin{tabular}{@{}lllll@{}}
    \toprule
    & \multicolumn{2}{c}{Cautious gated} & \multicolumn{2}{c}{Literal gated} \\
    \cmidrule(lr){2-3} \cmidrule(l){4-5}
    Wording distance & All plans & Among staged & All plans & Among staged \\
    \midrule
    Verbatim (``the switch to X went through'') & 25.0\% & 5/5 (100\%) & 100\% & 20/20 (100\%) \\
    Paraphrase & 30.0\% & 6/7 (85.7\%) & 75.0\% & 15/18 (83.3\%) \\
    Oblique (``went ahead with it'') & 25.0\% & 5/5 (100\%) & 85.0\% & 17/20 (85.0\%) \\
    Situational (change only implied) & 15.0\% & 3/6 (50.0\%) & 45.0\% & 9/17 (52.9\%) \\
    \bottomrule
  \end{tabular}
\end{table}

\subsection{Classifier per-class detail}
\label{app:classifier}

\begin{table}[H]
  \caption{Classifier precision and recall on 163 labeled natural statements (argmax labels; 95\% Wilson CIs). Overall accuracy is 69.9\% [62.5, 76.5].}
  \label{tab:perclass}
  \centering
  \footnotesize
  \begin{tabular}{@{}lrll@{}}
    \toprule
    Class & Support & Precision & Recall \\
    \midrule
    speculation & 42 & 92.5\%\ci{80.1, 97.4} & 88.1\%\ci{75.0, 94.8} \\
    completed event & 48 & 52.9\%\ci{42.5, 63.0} & 95.8\%\ci{86.0, 98.8} \\
    correction & 5 & 66.7\%\ci{20.8, 93.9} & 40.0\%\ci{11.8, 76.9} \\
    other & 68 & 87.9\%\ci{72.7, 95.2} & 42.6\%\ci{31.6, 54.5} \\
    \bottomrule
  \end{tabular}
\end{table}

\textbf{The ranking and argmax results answer different questions.} The 0.93 AUROC in Section~\ref{sec:setup} measures how well the classifier ranks statements that should update the profile. It uses the total probability assigned to commit classes. Table~\ref{tab:perclass} instead reports the single class with the highest probability.

\textbf{Evidence-type classification is better than direct truth questions.} A model can be confident that a user said ``I might move'' without being confident that the move occurred. This makes direct truth and storage questions unreliable in different cases. On the same statements, the evidence-type objective improves AUROC by $+0.24$\ci{0.16, 0.31} and $+0.31$\ci{0.23, 0.39} over the two alternatives.

\textbf{The main classifier error is over-commitment.} On a test against natural statements, its largest error is to label \emph{other} statements as completed events (35 of 68). These are not contamination cases. They show a bias toward the class that writes values immediately.

\textbf{Repeated runs are stable, except on corrections.} The classifier gives the same label in five repeated runs for 88.3\% of statements. Its majority label matches the reported run for 95.7\%. The probability scores also change little: the per-statement standard deviation averages 0.050, with a median of 0.016. As a result, AUROC remains between 0.933 and 0.950 across runs (mean 0.944). Most uncertainty comes from the correction class, which has only five examples. This small class widens the macro-F1 estimate to 0.665\ci{0.514, 0.774}. These cases also tend to be difficult for human reviewers: the blind human audit reproduces none of the five reference correction labels.

\subsection{Examples: a benchmark arc, the classifier prompt, and the field set}
\label{app:examples}

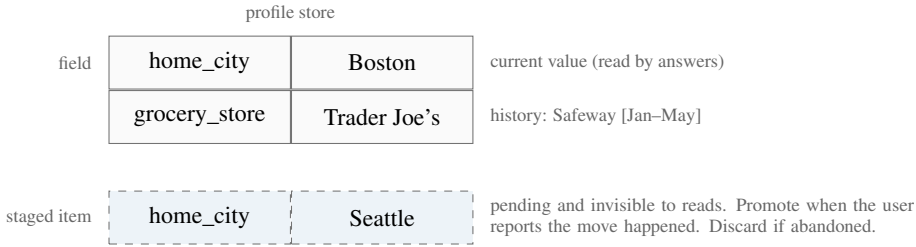
\begin{figure}[H]
  \centering
  \begin{tikzpicture}[font=\small,
      slot/.style={draw=black!55, fill=black!2, minimum height=0.7cm, minimum width=2.4cm, align=center},
      lab/.style={font=\scriptsize, text=black!60}]
    \node[lab] at (1.2,2.15) {profile store};
    \node[slot] (f1) at (0,1.5) {home\_city};
    \node[slot, right=0mm of f1] (v1) {Boston};
    \node[slot] (f2) at (0,0.78) {grocery\_store};
    \node[slot, right=0mm of f2] (v2) {Trader Joe's};
    \node[lab, left=1mm of f1] {field};
    \node[lab, right=1mm of v1] {current value (read by answers)};
    \node[lab, right=1mm of v2] {history: Safeway [Jan--May]};
    \node[slot, dashed, fill=litblue!8] (s1) at (0,-0.55) {home\_city};
    \node[slot, dashed, fill=litblue!8, right=0mm of s1] (sv) {Seattle};
    \node[lab, left=1mm of s1] {staged item};
    \node[lab, right=1mm of sv, text width=5.6cm]
      {pending and invisible to reads. Promote when the user reports the move happened. Discard if abandoned.};
  \end{tikzpicture}
  \caption{The profile store: each field is a slot holding one current value and a history. A staged item is a candidate value held outside the store with its two resolution conditions.}
  \label{fig:store}
\end{figure}

\paragraph{Example conversation.} This verbatim excerpt is from one of the 179 controlled-difficulty conversations. It uses the strongly hedged level, and the plan remains open. The correct current value is therefore ``Ghost 14''. The first session contains four more shopping exchanges. The second session starts eight days later, refers to the plan as still pending, and does not resolve it.

\begin{center}
\fbox{\begin{minipage}{0.94\linewidth}\small
\emph{Session 1 (2023/04/17):}\\
\textbf{User:} ``I run in Ghost 14 now, and the soles on my daily pair look pretty worn''\\
\textbf{Assistant:} ``If the outsole is smooth at the heel or forefoot, the midsole may also be tired. For a daily trainer replacement, start with what worked: fit, drop, cushion feel, and any hot spots. The Ghost line is steady and roomy; Pegasus tends to feel a bit snugger and more responsive. Check wear by putting the shoes on a table: if they tilt inward or the foam is creased hard on one side, replacement is fair. How many miles per week are you running, mostly roads or treadmill?''\\
\textbf{User (the hedge turn):} ``Here's where I'm stuck. I'm torn between staying with Ghost 14 and switching to Pegasus 39 for my next pair.''

\medskip
\emph{Session 2 (2023/04/25), opener:}\\
\textbf{User:} ``Oh, Pegasus 39 is still on the maybe list and I need a sale check. Anything good?''
\end{minipage}}
\end{center}

\paragraph{Example resolution conditions.} The staged item below is from a strong-hedge conversation whose user is torn between keeping ASICS Cumulus and switching to Saucony Ride 16. Both conditions are quoted verbatim from the run.

\begin{center}
\fbox{\begin{minipage}{0.94\linewidth}\footnotesize
\textbf{Promote when:} ``the user reports that they have actually switched their running
shoe model to the Saucony Ride 16, such as by buying them and using them as their current
running shoes. Evidence of merely considering, trying on, comparing, or planning to buy the
Saucony Ride 16 is not enough.''\\[2pt]
\textbf{Invalidate when:} ``the user reports that they decided against making the Saucony
Ride 16 their running shoe model, chose to keep using ASICS Cumulus or selected another
incompatible replacement shoe instead, or abandoned/cancelled the switch.''
\end{minipage}}
\end{center}

\paragraph{Classifier prompt.} The following text is quoted from the prompt file. It gives the gate question and defines the four classes.

\begin{center}
\fbox{\begin{minipage}{0.94\linewidth}\footnotesize\ttfamily
The gate question: does this statement license revising the\\
entity's canonical current state right now?\\
- speculative: asserts a possible, planned, intended, or otherwise\\
\hspace*{1.2em}unrealized future change; the asserted value is not yet true of the world.\\
- completed\_event: asserts that a state-changing event has occurred\\
\hspace*{1.2em}or that a new state now holds.\\
- explicit\_correction: asserts, with corrective intent, that a prior\\
\hspace*{1.2em}statement or stored belief about tracked state was wrong when it was made.\\
- other: non-revisionary content: additive observations, preferences,\\
\hspace*{1.2em}hedged recall of existing facts, schedules, questions.
\end{minipage}}
\end{center}

\paragraph{Field-set construction.} We fix the store fields before each run. Each held-out conversation has one field. Its attribute name comes from label metadata, not from an answer value. We create the field description from this name alone and attach the probe question that the model sees. We fixed the controlled-difficulty keys deterministically before generation. A validation check rejects any field that contains a label value. The field set is therefore value-free by construction.

\subsection{Product measurement detail}
\label{app:products}

We ran Mem0 and Zep Graphiti as released, using their own extraction, storage, and retrieval, and configured them to use the same pinned \texttt{gpt-5.5-2026-04-23} model as our systems. Each of the 147 held-out conversations was ingested session by session. After every session, we asked the product for the user's current value of the tracked field, and the fixed equivalence judge of Section~\ref{sec:setup} scored the answer against the ground-truth current value. A plan counts as exposed if any answer before the plan's outcome asserts it as the user's current state. Answers worded as plans do not count. Under this reading, Mem0 asserted never-true plans as current for 31 of 88 at-risk plans (35.2\%) and Graphiti for 24 of 88 (27.3\%). These are whole-product measurements, so differences against our component-level systems cannot be attributed to any single stage. Library versions and configuration are recorded in the released code.


\end{document}